\documentclass[conference]{IEEEtran}
\usepackage{times}

\usepackage[numbers]{natbib}
\usepackage{multicol}
\usepackage[bookmarks=true]{hyperref}

\usepackage{booktabs,multirow,array}
\newcommand{\thickhline}{\noalign{\hrule height 1.1pt}}
\newcommand{\method}{Physics-Guided Residual Dynamics\xspace}
\newcommand{\methodshort}{PGRD\xspace}
\usepackage[table]{xcolor}
\definecolor{Gray}{gray}{0.9}
\newcommand{\figref}[1]{Fig.~\ref{#1}}
\newcommand{\secref}[1]{Sec.~\ref{#1}}
\newcommand{\appref}[1]{App.~\ref{#1}}
\newcommand{\tabref}[1]{Tab.~\ref{#1}}

\usepackage{xspace}
\usepackage{graphicx}
\usepackage{amsmath}
\usepackage{amssymb}
\let\labelindent\relax
\usepackage{enumitem}
\usepackage{caption}
\usepackage{xcolor}
\usepackage{stfloats}

\usepackage{etoolbox}
\AtBeginDocument{%
  \setlength{\abovedisplayskip}{5pt}%
  \setlength{\belowdisplayskip}{5pt}%
  \setlength{\abovedisplayshortskip}{0pt}%
  \setlength{\belowdisplayshortskip}{0pt}%
}

\newcommand{\maketitlesupplementary}{
\twocolumn[
\begin{center}
\vspace{1em}
{\LARGE\bfseries Appendix}
\vspace{1em}
\end{center}
]
}

\begin{document}

\title{Learning Physics-Guided Residual Dynamics for Deformable Object Simulation}

\author{Author Names Omitted for Anonymous Review. Paper-ID 54}

\author{
  Shivansh Patel$^{1}$ \quad
  Kaifeng Zhang$^{2*}$ \quad
  Sanjay Pokkali$^{1*}$ \quad
  Svetlana Lazebnik$^{1}$ \quad
  Yunzhu Li$^{2}$ \\[0.5em]
  \small{$^{1}$UIUC \quad
  $^{2}$Columbia University}
}

\maketitle

\begin{abstract}

Simulating deformable objects is essential for a wide range of robotic manipulation applications, yet accurately predicting their dynamics remains challenging. We propose \method (\methodshort), a hybrid simulation framework that combines the advantages of physics-based and learning-based approaches. Specifically, \methodshort combines an optimizable spring–mass simulator as a backbone with a learned neural network that predicts residual corrections to the physics-based predictions. We adopt a velocity-based formulation to ensure stable simulation and a sliding-window transformer architecture to capture temporal dependencies. We show that \methodshort produces more accurate results than both purely physics-based and learning-based methods on a set of diverse real-world deformable objects. We further demonstrate the utility of \methodshort in two applications: manipulation planning via Model Predictive Control, including a language-conditioned setting with a generated goal image; and interactive simulation via action-conditioned video prediction by 3D Gaussian Splatting. Project page: \url{https://pgrd-robot.github.io/}
\end{abstract}

\IEEEpeerreviewmaketitle

\begin{figure*}[b]
  \centering
  \includegraphics[width=\textwidth]{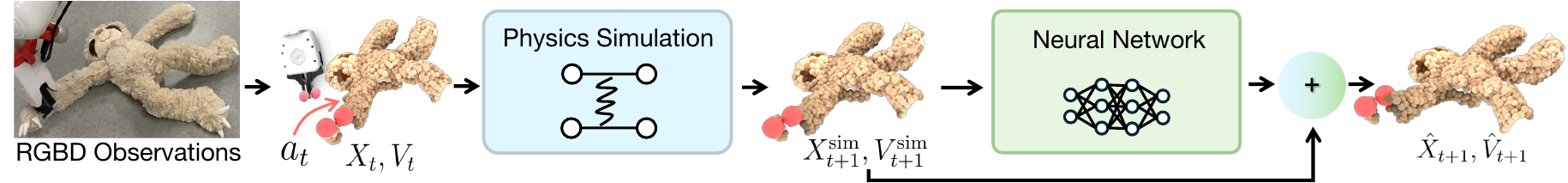}
  \vspace{-0.4cm}  %
  \caption{\small \textbf{\method Overview.} Given RGBD observations, we extract surface points to instantiate the simulation. Red dots indicate the point where gripper held the object. The framework rolls out an optimized spring-mass physics backbone from the current object state $X_t, V_t$ and robot actions $a_t$ to produce simulation prediction $X_{t+1}^{\text{sim}}, V_{t+1}^{\text{sim}}$. Subsequently, our residual dynamics network predicts per particle residual velocities, which are added to the simulator velocities and time integrated to obtain the final positions $\hat{X}_{t+1}, \hat{V}_{t+1}$.}
  \label{fig:method}
  \vspace{-0.4cm}
\end{figure*}

\section{Introduction}
\label{sec:intro}
 Accurate simulation of deformable objects is a persistent challenge in robotics and computer vision. Such objects can change shape significantly under external forces, undergoing stretching, bending, crumpling, and twisting. Simulation difficulty stems from material properties, such as heterogeneous elasticity and damping, but is also complicated by factors, such as self-collisions and friction, that come into play during contact-rich interactions.

The literature contains two dominant paradigms for deformable object simulation: physics-based methods~\cite{baraff1998,stomakhin2013material,muller2007position} and learning-based methods~\cite{yan2020self,chen2022comphy,kipf2019contrastive} (see Section \ref{sec:rel_works} for a more detailed survey of related works). Physics-based methods use equations to describe how materials deform and respond to forces, yielding physically plausible and interpretable simulations. However, these methods require precise material parameters that are difficult to obtain in practice~\cite{lloyd2007identification,yang2025differentiable,murthy2020gradsim}. The complex mathematics of these simulators typically restricts practitioners to gradient-free optimization~\cite{lloyd2007identification}, which can tune only a handful of parameters before becoming intractable~\cite{yang2025differentiable,zhu2025deformable}. Even with optimal parameters, physics-based models are inherently limited by discretization (coarse meshes miss fine-scale deformations~\cite{tan2021hybrid,madier2023fundamentals}) and modeling assumptions (linear elasticity fails under large strains~\cite{webster2010design,mihai2017characterize}).
Learning-based methods sidestep these limitations by fitting observations that are challenging to model analytically. However, purely data-driven models are data-hungry~\cite{yu2024learning} and suffer from poor generalization to unseen scenarios~\cite{di2022physically,vasiliauskaite2024generalization}. They often lack the physical consistency required for real-world deployment~\cite{di2022physically} and, without the inductive bias provided by physical priors, they may learn spurious correlations~\cite{wang2022causal}.

To combine the interpretability and generalizability of physics-based models with the flexibility of learning-based models, we propose a hybrid simulation framework of \method (\methodshort). As shown in \figref{fig:method} and explained in Section \ref{sec:method}, \methodshort uses a spring-mass simulator as its physics backbone and a neural network to predict residual corrections to that model. \methodshort follows a two-stage training procedure: first, we optimize the simulator's parameters to match real-world observations; and second, we train the neural network to compensate for the discrepancies between the physics model and the data. Note that making this two-stage approach work is nontrivial, as directly injecting learned position corrections into simulated states can destabilize the dynamics and cause error accumulation over time. Motivated by impulse-based simulation that evolves states through velocity updates rather than position~\cite{mirtich1995impulse,stewart2000implicit,stewart2000implicit}, we predict residual velocities and integrate them forward in time, yielding position corrections that respect the underlying dynamics. This velocity-based formulation enables smooth, stable training and simulation. To model temporal dependencies, we incorporate a sliding-window transformer that refines velocity corrections across multiple timesteps. This temporal history enables the network to capture dynamic phenomena such as momentum, while a gating mechanism ensures stable training.

Compared to purely physics-based or learning-based simulation methods, \methodshort 
offers several advantages. First, the physics backbone provides priors that can improve generalization to new scenarios. Second, it is expected to require less training data than purely learning-based approaches, as the network only needs to predict corrections rather than the entire dynamics from scratch. Third, it maintains computational efficiency while achieving higher accuracy than physics-based methods alone. To further reduce compounding errors over long horizons, we train \methodshort with multi-step rollouts: during training, the model is recursively applied to its own predicted states rather than being reset to ground truth at every step. This exposes the residual model to the same distribution shift encountered at test time and is important for stable long-horizon simulation.

In Section \ref{sec:experiments}, we validate \methodshort on a diverse set of real-world deformable objects, including rope, paper, soft toys, flag, and duster. 
Across these objects, we evaluate simulation quality against tracked 3D point trajectories from real-world executions, covering both prehensile and non-prehensile manipulation.
Our experiments demonstrate that \methodshort consistently outperforms both purely physics-based and purely learning-based approaches in tracking accuracy. Notably, only \methodshort performs well on the duster, which is a heterogeneous object with a rigid stem and soft feathers. Here, uniform material assumptions in most physics-based deformable simulators fail and learning-based methods struggle with the complexity. Besides standard tracking evaluation, we also show the advantage of \methodshort for action-conditioned video prediction, which we perform by attaching 3D Gaussians to the simulated particles.

Beyond evaluation, we demonstrate two applications of \methodshort in Section \ref{sec:applications}. First, \methodshort can serve as a forward model for \textit{manipulation planning} via Model Predictive Path Integral (MPPI) control, where candidate action sequences are rolled out and ranked by Chamfer Distance to a target configuration. We demonstrate this on challenging tasks such as cable rerouting through a narrow slot, where \methodshort succeeds in 8 out of 10 trials compared to 2 out of 10 for the tuned spring-mass baseline. We further extend planning to a language-conditioned setting, where a goal image generated from a language command replaces the need for a pre-collected target point cloud. Second, \methodshort enables \textit{interactive photorealistic simulation} in which users issue manipulation commands and the predicted particle states drive 3D Gaussians to render photorealistic views of the resulting deformation.

In summary, our contributions are as follows. (1) We propose \method, a hybrid simulation framework that combines an optimizable spring-mass model with learned residual corrections to accurately model deformable object dynamics. (2) We introduce a two-stage training procedure that first optimizes physics parameters using black-box optimization and then trains a neural network to predict residual corrections. (3) We conduct extensive experiments on diverse real-world deformable objects, demonstrating that our approach achieves superior performance compared to existing physics-based and learning-based methods.

\section{Related Works}
\label{sec:rel_works}

\noindent\textbf{Physics-Based Simulation for Deformable Objects.}
Traditional physics-based simulation methods rely on analytical models to discretize deformable objects and numerical solvers for the equations of motion of the models. For deformable objects, spring-mass models~\cite{baraff1998,liu2013fast,liang2019differentiable} represent one of the most intuitive and efficient approaches, where objects are modeled as networks of point masses interconnected by springs. Consequently, this representation has been extensively employed to model the dynamics of diverse deformable objects in both computer graphics and robotics~\cite{baraff1998,liu2013fast,liang2019differentiable,zhong2023improving,lloyd2007identification,zhong2024reconstruction,jiang2025phystwin}. Our work leverages the spring-mass model as a backbone precisely because of its efficiency and interpretability. While more complex approaches like Finite Element Methods (FEM)~\cite{sifakis2012fem,muller2004interactive}, Position-Based Dynamics (PBD)~\cite{muller2007position,macklin2014unified}, and Material Point Methods (MPM)~\cite{stomakhin2013material,jiang2015affine} offer higher physical fidelity, they incur high computational costs and complexity. We include MPM as a baseline in our experiments to validate this tradeoff.

Despite their physical foundations, the above-mentioned methods require precise material parameters that are difficult to obtain in practice. A recent line of work addresses this challenge by reformulating physics simulators to be compatible with automatic differentiation~\cite{liang2019differentiable,um2020solver,holl2020learning,belbute2020combining,du2021diffpd,geilinger2020add,heiden2021disect,jatavallabhula2021gradsim,qiao2021differentiable,rojas2021differentiable,ma2022risp,liu2024softmac,gao2024sim,jiang2025phystwin}. These methods identify optimal physics parameters by backpropagating through the simulation process, solving inverse problems to find material properties that best match observed data. While such parameter identification can improve simulation accuracy, it imposes a strong requirement the simulator must be fully differentiable. This is often impractical in contact-rich robotics tasks where discontinuities from collisions~\cite{beker2025smooth,lidec2024end,zhong2023improving}, friction mode transitions~\cite{lidec2024end,kim2025online,halm2021set}, and non-smooth contact geometry~\cite{werling2021fast,ye2025efficient,lidec2024end} lead to exploding or vanishing gradients. In practice, even state-of-the-art differentiable simulation methods suffer from these fundamental limitations. We compare against PhysTwin~\cite{jiang2025phystwin}, a recent state-of-the-art differentiable spring-mass approach, across all of our tasks, and show that it additionally struggles on heterogeneous deformable objects. Our method does not require the simulator to be differentiable, enabling broader applicability to realistic manipulation scenarios.

\vspace{2mm}
\noindent\textbf{Learning-Based Simulation for Deformable Objects.} Learning has emerged as a powerful alternative to traditional physics-based simulation, particularly excelling in scenarios requiring real-time performance with complex nonlinear dynamics~\cite{yan2020self,chen2022comphy,kipf2019contrastive,zhou2024dino,xue2023neural,bauer2024doughnet,ai2025review,huang2025particleformer,chua2018deep,evans2022context,ma2023learning,wu2019learning,xu2019densephysnet,luiten2024dynamic,huang2026pointworld}. These approaches leverage data-driven models, typically neural networks, to learn deformation patterns directly from observation data, bypassing the need for explicit material parameters or complex numerical solvers~\cite{li2018learning,mrowca2018flexible}.

Graph Neural Networks (GNNs) have become prominent in this domain due to their natural ability to represent meshes as graphs. GNNs excel at capturing long-range interactions and complex dependencies through message passing~\cite{zhang2024adaptigraph,zhang2024dynamic,sanchez2020learning,pfaff2020learning,li2018learning,wang2022offline,li2019propagation}, allowing them to effectively simulate internal and external forces causing deformation~\cite{battaglia2016interaction,kipf2018neural}. 
Notably, GBND~\cite{zhang2024dynamic} establishes a topology over sparse particles, enabling the GNN to efficiently learn and propagate the dynamics across the object. We compare against GBND in our experiments. However, GNN-based methods face significant challenges: the effectiveness of message passing is sensitive to graph structure and is vulnerable to partial observations, while insufficient steps fail to capture global information and excessive steps cause oversmoothing~\cite{wang2022graph,attali2024rewiring}.

Recent learning-based methods have explored architectural improvements, including transformers for handling dense particles~\cite{shao2022transformer,whitney2024modeling}, recurrent neural networks for temporal consistency~\cite{ma2022learning}, and attention mechanisms for improved long-range dependency modeling~\cite{sanchez2020learning}. These methods have demonstrated success across diverse materials, including cloth~\cite{liang2019differentiable,lin2022learning,pfaff2020learning}, fluids~\cite{sanchez2020learning,li2018learning}, plasticine~\cite{huang2021plasticinelab,shi2024robocraft,shi2023robocook}, and granular materials~\cite{li2018learning}. A recent state-of-the-art method is Particle-Grid Neural Dynamics~\cite{zhang2025particle} (PGND), which combines particle representations with spatial grids to learn dynamics while maintaining spatial continuity, and serves as a strong baseline in our experiments. On the down side, approaches like PGND require substantial training data and may struggle with generalization to unseen scenarios or material properties that are significantly different from the training distributions. Our work addresses these limitations by grounding the neural network in physical priors, allowing it to focus on learning only residual corrections rather than the full dynamics from scratch. 

\vspace{2mm}
\noindent\textbf{Residual Dynamics.} The idea of residual learning has been previously employed to model the dynamics of robots~\cite{golemo2018sim,gruenstein2021residual,bauersfeld2021neurobem,dong2026learning}. In these works, the robot's model is known, and the residuals primarily reflect simple modeling errors, such as PID gains or backlash. These discrepancies are relatively easy to capture because they often manifest as consistent deviations from the nominal model. In contrast, we model the dynamics of deformable objects, where the residual must compensate for complex, high-dimensional phenomena, including nonlinear stiffness, spatially varying material properties, and contact-rich interactions. The most closely related work in spirit is~\cite{ajay2018augmenting}, which models the environment's residual dynamics rather than the robot's. However, it is limited to low-dimensional toy problems such as planar pushing, and its simple architecture does not scale to the high-dimensional particle cloud representations required for deformable object manipulation, making a direct comparison infeasible. To the best of our knowledge, \methodshort is the first work to extend the residual paradigm to modeling high-dimensional, complex states of deformable objects.

\section{Method}
\label{sec:method}

This section presents the details of our \method method, which is illustrated in \figref{fig:method}. We first explain our physics backbone, followed by the residual dynamics framework.

\subsection{Physics Backbone}
Given RGBD observations, we extract surface points to instantiate the simulation.
We represent deformable objects using a spring-mass model, wherein the object is discretized into a graph structure consisting of point masses (nodes) interconnected by springs (edges). Each node $i$ has a position $\mathbf{x}_i \in \mathbb{R}^3$ and velocity $\mathbf{v}_i \in \mathbb{R}^3$ that evolve over time according to Newtonian mechanics with a semi-implicit Euler solver. This representation provides computational efficiency and straightforward implementation while capturing essential elastic deformation behaviors. To model manipulation, we apply constraints to particles near the gripper: for prehensile manipulation, they move rigidly with the gripper, whereas for non-prehensile manipulation, the gripper acts as a spherical collision shape that pushes penetrating particles to its surface.

The fidelity of the spring-mass simulator depends critically on five physical parameters collectively denoted as $\theta$: 
1) \textit{Stiffness}, which defines the elastic resistance of springs; 
2) \textit{Damping}, which controls energy dissipation during motion; 
3) \textit{Threshold}, which determines the maximum distance for creating springs between nodes; 
4) \textit{Max springs per node}, which limits the connectivity of each mass point for simulation stability; and 
5) \textit{Ground friction}, which governs contact interactions with the environment. 
These parameters collectively encode the material properties and environmental conditions that govern the dynamics of the simulated object.

Given a batch of trajectories sampled from the dataset, each with initial state $(\mathbf{X}_0,\mathbf{V}_0)$ and action sequences $\mathcal{A} = \{a_t\}_{t=0}^{T-1}$, we roll out the simulator for $T$ time steps with actions $\{a\}_{t=0}^{T-1}$ under candidate parameters $\theta$ to obtain predicted point cloud positions $\{\hat{\mathbf{X}}_t\}_{t=1}^T$. At each time step $t$, we compute a pointwise mean squared error between predicted and ground truth configurations. We average this error over the trajectory horizon to obtain the optimization objective. We minimize this objective using Covariance Matrix Adaptation Evolution Strategy (CMA-ES)~\cite{hansen2001completely}, which is particularly effective for our problem as it optimizes parameters without requiring gradients. We use CMA-ES because contacts and collisions cause sudden changes in forces, preventing reliable gradient computation in the simulator.

For volumetric objects such as soft toys, modeling only the surface shell often leads to unrealistic collapse, as the lack of internal support points causes the object to flatten under gravity or compression. 
To address this, we augment the spring-mass model with internal points that provide structural volume. 
Starting from our multi-view observations, we first reconstruct the complete surface point cloud using RaySt3R~\cite{rayst3r} and extract a watertight mesh via marching cubes. 
We then uniformly sample particles within this mesh to populate the interior, ensuring the spring-mass model preserves the object's volume and structural integrity during simulation.

\subsection{Residual Dynamics Framework}
While our optimized spring-mass backbone provides a physically grounded baseline, it inherently struggles to capture complex behaviors such as nonlinear stiffness, non-uniform contact friction, and heterogeneous material properties. 
Instead of attempting to model these intricacies analytically, we introduce a learned residual module to predict the discrepancy between the physics model and observations. 

Formally, at each time step \(t\) and step size $dt$, we use the physics backbone to generate an immediate prediction \((X^{\text{sim}}_{t+1}, V^{\text{sim}}_{t+1})\).
We then employ a neural network to predict a per-particle residual velocity \(\Delta V^{\text{residual}}_{t+1}\) based on the current simulator state and history.
An alternative design would be to predict residual positions directly. However, we found this formulation to be unstable in practice: directly correcting positions led to occasional simulation blow-ups, where object state becomes NaNs. 
Intuitively, directly predicting positions can
create a mismatch between the corrected positions and the
velocities, destabilizing the simulation.
We therefore adopt a velocity-based residual formulation, which ensures smooth integration and avoids the instability often associated with direct position corrections.
The final velocities are obtained by adding the learned residuals to the simulator's predictions:
\begin{equation*}
\hat{V}_{t+1} = V^{\text{sim}}_{t+1} + \Delta V^{\text{residual}}_{t+1}.
\end{equation*}
These corrected velocities are then time-integrated to obtain the final position updates:
\begin{equation*}
\hat{X}_{t+1} = X^{\text{sim}}_{t+1} + \Delta V^{\text{residual}}_{t+1} \cdot dt.
\end{equation*}
For particles that are rigidly held by grippers, we enforce a boundary condition by zeroing out their residuals. This formulation effectively bridges the gap between the optimized spring-mass model and the observed real-world dynamics.

We train the residual network with a supervised objective that minimizes the mean squared error between predicted and ground truth particle positions over a rollout:
\begin{equation}
\mathcal{L}_{\text{res}}
=
\frac{1}{T}
\sum_{t=1}^{T}
\frac{1}{N}
\left\|
\hat{X}_t - X_t^{*}
\right\|_2^2,
\end{equation}
where \(\hat{X}_t\) denotes the predicted particle positions at rollout step \(t\), \(X_t^{*}\) denotes the corresponding ground truth positions, and \(N\) is the number of particles. To improve robustness to error accumulation over long horizons, we apply this loss over multi-step rollouts rather than a single step. During training, we do not reset the simulator to the ground truth state at every step. Instead, we feed the hybrid simulator's predicted state at time \(t\) back as the input for time \(t+1\). This exposes the network to its own past predictions and allows it to learn how to recover from drift.

\subsection{Network Architecture}
\label{sec:network_arch}

\begin{figure}[h]
  \centering
  \vspace{-0.1cm}
  \includegraphics[width=\linewidth]{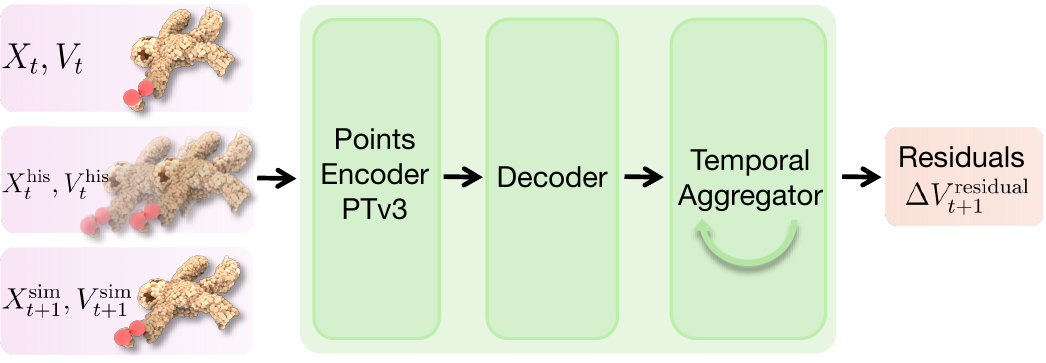}
  \vspace{-0.4cm}  %
    \caption{\small \textbf{Network Architecture.} The points encoder extracts spatiotemporal features, which the decoder projects to estimate initial residual velocity. Finally, the temporal aggregator refines these estimates using a recurrent history to output the final residuals $\Delta V^{\text{residual}}_{t+1}$.}
  \label{fig:network}
  \vspace{-0.2cm}
\end{figure}

To parameterize the residual velocity, we design a network consisting of three primary components: encoder, decoder, and temporal aggregator. The network architecture is shown in \figref{fig:network}. While individual components draw on established design choices, our contribution lies in their integration into a single residual-dynamics architecture. We ablate the architecture components in \appref{sec:ablation_appendix} and provide the complete hyperparameters for the network components in \appref{app:network_details}.

\vspace{2mm}
\noindent\textbf{Encoder.} 
We employ a Point Transformer V3 (PTv3)~\cite{wu2024point} architecture to extract per-particle features. At time step \( t \), we construct input features by concatenating the current state \( (X_{t}, V_{t}) \), history from previous time steps \( (X^{\text{his}}_{t}, V^{\text{his}}_{t}) \), and the simulator's immediate predictions \( (X^{\text{sim}}_{t+1}, V^{\text{sim}}_{t+1}) \). The encoder processes these inputs through multiple layers of patch-based self-attention, where each particle attends to others within its local neighborhood. We demonstrate experimentally that the attention mechanism computes position-dependent features for each particle based on its local geometric configuration.

\vspace{2mm}
\noindent\textbf{Decoder.} 
To decode per-particle residuals, we utilize a Neural Radiance Field (NeRF) style architecture. We apply Fourier positional encoding to the particle positions and concatenate the resulting embeddings with the features extracted by the encoder. These concatenated features are passed through an MLP decoder to produce an initial velocity correction estimate $\Delta V^{\text{initial}}_{t+1}$. Empirically, we found that including the decoder improves performance compared with directly using PTv3 outputs as velocity estimates, as discussed in the decoder ablation in \appref{sec:ablation_decoder}.

\vspace{2mm}
\noindent\textbf{Temporal Aggregator.} 
To incorporate dynamics from the previous predictions, we refine the base predictions using a sliding-window transformer. At each step $t$, we update a temporal buffer with the current initial correction $\Delta V^{\text{initial}}_{t+1}$. If the buffer contains fewer than $W$ frames, we pad it by replicating the most recent frame. 
The sequence is projected to a latent embedding, augmented with sinusoidal positional encodings, and processed by the transformer. We extract the output feature vector $\mathbf{h}_t$ corresponding to the final time step and apply two parallel learned projections:
\begin{equation*}
\boldsymbol{\delta}_t = \tanh(\text{Proj}_\delta(\mathbf{h}_t)), \quad g_t = \sigma(\text{Proj}_g(\mathbf{h}_t)),
\end{equation*}
where $\text{Proj}$ denotes a linear layer, $\tanh$ produces a bounded temporal offset $\boldsymbol{\delta}_t$, and the sigmoid function $\sigma$ generates gating weights $g_t$. 
The final residual velocity is computed as:
\begin{equation*}
\Delta V^{\text{residual}}_{t+1} = 0.1 \cdot [(1 - g_t) \odot \Delta V^{\text{initial}}_{t+1} + g_t \odot \boldsymbol{\delta}_t],
\end{equation*}
where $\odot$ denotes element-wise multiplication. This gating mechanism allows the network to adaptively blend the local base prediction with the temporally refined correction based on the transformer's context. This design is similar in spirit to gated recurrent models~\cite{hochreiter1997long,cho2014learning}, in that it uses temporal context to decide how much of the current prediction to preserve and how much to replace with a history-based update.

\section{Experiments}
\label{sec:experiments}

\subsection{Data Collection}
\begin{figure}[h]
  \centering
  \includegraphics[width=\linewidth]{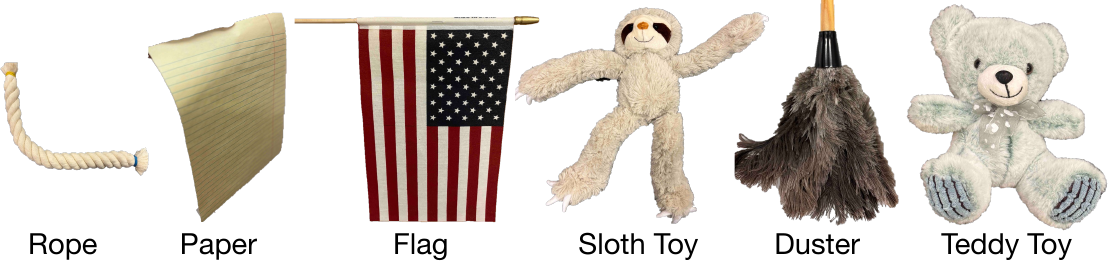}
  \vspace{-0.6cm} 
  \caption{\small \textbf{Experimental Objects} (see text).}
  \label{fig:objects}
  \vspace{-3mm}
\end{figure}

We evaluate \methodshort on six objects shown in \figref{fig:objects} spanning a range of shapes, material properties, and manipulation types.

\begin{enumerate}[label=\textbf{\alph*)}, leftmargin=2em, itemsep=0.25em]
  \item \textbf{Rope.} A mostly one-dimensional object grasped at one end and manipulated by dragging across the table or lifting and lowering motions.
  \item \textbf{Paper.} A thin two-dimensional sheet grasped at the top and manipulated by suspending it in the air while performing waving motions, exhibiting bending and twisting.
  \item \textbf{Flag.} A cloth flag, demonstrating large area deformation dynamics that are challenging to model due to the waving motion. We chose it because its top stick prevents the cloth's surface from collapsing. We model the stick as a rigid constraint, assuming that all points along the stick follow the gripper trajectory.
  \item \textbf{Sloth Toy.} A three-dimensional volumetric object with long, highly flexible limbs that are challenging to model.
  \item \textbf{Duster.} A heterogeneous object with a rigid stem and highly deformable feathers. This is especially challenging because it requires different dynamics predictions for the rigid stem compared to the flexible feathers.
  \item \textbf{Teddy Toy.} A volumetric object with higher rigidity than the soft toy. We deform it by poking rather than holding it, representing non-prehensile manipulation.
\end{enumerate}

We collect data using UFACTORY xArm 7 following the procedure from~\cite{zhang2025particle}. Four Intel RealSense D455 cameras capture synchronized multi-view observations while the object is manipulated. We segment the object with Grounded SAM 2~\cite{ren2024grounded} and track image points with CoTracker~\cite{karaev2025cotracker3}. These 2D tracks are lifted to 3D using depth from all camera views, and an iterative rollout procedure is used to extract persistent correspondences across frames. The resulting dataset consists of temporally consistent 3D point cloud trajectories of the object together with the robot gripper positions over time, obtained from robot proprioception and used as the action sequence. Additional details are provided in \appref{app:data_coll}.

For training, we use 100 episodes, and for validation, we use 20 episodes. Each episode spans 4 seconds, but consecutive episodes are generated with a temporal stride of 2 seconds, so neighboring episodes overlap by 2 seconds. As a result, the 100 training episodes correspond to 202 seconds of unique data, or approximately 3.37 minutes, while the 20 validation episodes correspond to 42 seconds, or 0.70 minutes. In total, the dataset contains 120 episodes extracted from 244 seconds of interaction, which is approximately 4.07 minutes of recorded data. The entire data processing is offline and happens before training. We train \methodshort with 5-step rollout, but evaluate it over the full evaluation episode length of 37 steps. Hence, this evaluation paradigm tests generalization capability over horizons not seen during training.

\begin{table*}[t]
    \centering
    \resizebox{\textwidth}{!}{%
    \begin{tabular}{@{} l l !{\vrule width 1pt} c c c c c c @{}}
        \thickhline
        Method & Metric & Rope & Paper & Flag & Sloth Toy & Duster & Teddy Toy\\
        \thickhline
        
        Spring-Mass~\cite{jiang2025phystwin} & \multirow{6}{*}{MDE $\downarrow$}
        & $4.4_{\pm 2.0}$
        & $2.3_{\pm 2.0}$
        & $5.7_{\pm 2.5}$
        & $3.2_{\pm 0.8}$
        & $3.8_{\pm 2.0}$
        & $5.8_{\pm 3.4}$ \\

        MPM~\cite{stomakhin2013material} & 
        & $7.3_{\pm 2.5}$
        & $15.5_{\pm 4.2}$
        & $23.1_{\pm 7.0}$
        & $7.4_{\pm 1.8}$
        & $6.0_{\pm 2.6}$
        & -- \\

        GBND~\cite{zhang2024dynamic} & 
        & $5.5_{\pm 1.7}$
        & $3.0_{\pm 1.4}$
        & $30.9_{\pm 6.2}$
        & $7.7_{\pm 2.6}$
        & $5.1_{\pm 2.2}$
        & $1.5_{\pm 0.4}$ \\

        PGND~\cite{zhang2025particle} & 
        & $3.3_{\pm 1.8}$
        & $2.1_{\pm 0.5}$
        & $3.2_{\pm 1.4}$
        & $4.0_{\pm 1.3}$
        & $3.8_{\pm 0.1}$
        & $1.6_{\pm 1.3}$ \\

        Diff. Spring-Mass~\cite{jiang2025phystwin} &
        & $3.5_{\pm 1.7}$
        & $1.9_{\pm 1.6}$
        & $4.6_{\pm 2.2}$
        & $2.9_{\pm 0.7}$
        & $3.5_{\pm 1.6}$
        & $4.6_{\pm 2.7}$ \\

        Ours & 
        & $\mathbf{2.6_{\pm 1.31}}$
        & $\mathbf{1.7_{\pm 0.7}}$ 
        & $\mathbf{2.8_{\pm 1.2}}$
        & $\mathbf{2.7_{\pm 0.4}}$ 
        & $\mathbf{2.9_{\pm 1.8}}$ 
        & $\mathbf{1.3_{\pm 0.3}}$\\
        \thickhline
        
        Spring-Mass~\cite{jiang2025phystwin} & \multirow{6}{*}{CD $\downarrow$}
        & $4.5_{\pm 2.4}$ 
        & $1.7_{\pm 1.2}$ 
        & $9.1_{\pm 4.7}$ 
        & $2.9_{\pm 0.7}$ 
        & $4.7_{\pm 2.5}$ 
        & $5.1_{\pm 3.0}$ \\

        MPM~\cite{stomakhin2013material} & 
        & $6.1_{\pm 2.4}$ 
        & $14.8_{\pm 6.5}$ 
        & $23.6_{\pm 10.4}$  
        & $7.1_{\pm 1.7}$ 
        & $5.2_{\pm 1.7}$ 
        & -- \\

        GBND~\cite{zhang2024dynamic} & 
        & $6.6_{\pm 2.4}$ 
        & $5.0_{\pm 1.7}$ 
        & $49.1_{\pm 13.0}$ 
        & $6.5_{\pm 1.8}$ 
        & $6.3_{\pm 2.8}$ 
        & $2.7_{\pm 0.3}$ \\

        PGND~\cite{zhang2025particle} & 
        & $2.7_{\pm 1.4}$ 
        & $2.2_{\pm 0.9}$ 
        & $4.3_{\pm 2.4}$ 
        & $3.2_{\pm 0.8}$ 
        & $4.0_{\pm 0.1}$ 
        & $1.6_{\pm 1.0}$ \\

        Diff. Spring-Mass~\cite{jiang2025phystwin} &
        & $3.4_{\pm 1.8}$
        & $1.5_{\pm 1.0}$
        & $6.8_{\pm 3.6}$
        & $2.8_{\pm 0.7}$
        & $4.3_{\pm 2.1}$
        & $3.8_{\pm 2.3}$ \\

        Ours & 
        & $\mathbf{2.3_{\pm 1.3}}$
        & $\mathbf{1.3_{\pm 0.6}}$ 
        & $\mathbf{4.3_{\pm 2.2}}$
        & $\mathbf{2.6_{\pm 0.3}}$ 
        & $\mathbf{3.8_{\pm 2.5}}$ 
        & $\mathbf{1.3_{\pm 0.2}}$ \\
        \thickhline
        
        Spring-Mass~\cite{jiang2025phystwin} & \multirow{6}{*}{EMD $\downarrow$}
        & $2.4_{\pm 1.3}$ 
        & $1.8_{\pm 1.2}$ 
        & $4.9_{\pm 2.6}$ 
        & $1.5_{\pm 0.4}$ 
        & $1.9_{\pm 1.2}$ 
        & $2.9_{\pm 1.8}$ \\

        MPM~\cite{stomakhin2013material} & 
        & $3.9_{\pm 1.6}$ 
        & $11.4_{\pm 5.5}$ 
        & $19.4_{\pm 7.8}$ 
        & $3.9_{\pm 0.9}$ 
        & $2.8_{\pm 1.2}$ 
        & -- \\

        GBND~\cite{zhang2024dynamic} & 
        & $3.0_{\pm 1.3}$ 
        & $2.1_{\pm 0.9}$ 
        & $24.5_{\pm 6.3}$ 
        & $3.1_{\pm 1.1}$ 
        & $2.8_{\pm 1.4}$ 
        & $0.8_{\pm 0.2}$\\

        PGND~\cite{zhang2025particle} & 
        & $1.4_{\pm 0.6}$ 
        & $1.5_{\pm 0.5}$ 
        & $2.3_{\pm 1.3}$ 
        & $1.6_{\pm 0.4}$ 
        & $1.6_{\pm 0.4}$ 
        & $0.8_{\pm 0.6}$ \\

        Diff. Spring-Mass~\cite{jiang2025phystwin} &
        & $1.9_{\pm 1.0}$
        & $1.6_{\pm 1.0}$
        & $3.9_{\pm 2.1}$
        & $1.6_{\pm 0.4}$
        & $1.6_{\pm 1.0}$
        & $2.3_{\pm 1.3}$ \\

        Ours & 
        & $\mathbf{1.2_{\pm 0.6}}$
        & $\mathbf{1.4_{\pm 0.7}}$ 
        & $\mathbf{2.2_{\pm 1.2}}$
        & $\mathbf{1.4_{\pm 0.2}}$ 
        & $\mathbf{1.4_{\pm 1.1}}$ 
        & $\mathbf{0.6_{\pm 0.1}}$ \\
        \thickhline
    \end{tabular}
    }
    \caption{\small \textbf{Tracking accuracy across diverse objects.} \method achieves the lowest tracking error across all objects and metrics, outperforming both physics-based methods and learning-based approaches.}
    \label{tab:results-tracking}
\end{table*}

\subsection{Evaluation Tasks and Metrics} \label{sec:tasks}
We evaluate \methodshort in two settings. First, we measure how accurately it predicts object motion in 3D using tracking metrics. Second, we evaluate whether the same predicted trajectories produce visually accurate future observations in image space. In both settings, performance is measured over complete validation episode rollouts rather than isolated timesteps. For the visual-space evaluation, we convert the predicted 3D trajectories into 2D video frames using 3D Gaussian Splatting. We initialize a 3DGS representation from the first frame and then update the Gaussians using the predicted particle motion while keeping appearance parameters fixed. This yields action-conditioned video predictions that can be compared directly against ground-truth frames using visual metrics.

\noindent\textbf{Tracking Metrics.} We assess rollout accuracy in 3D using three metrics:
\begin{itemize}[leftmargin=*,labelsep=5pt]
    \item \textbf{Mean Distance Error (MDE)} measures the average distance (in cm) between corresponding points in
    predicted and ground truth configurations, providing a direct assessment of positional accuracy.
    \item \textbf{Chamfer Distance (CD)} computes the bidirectional nearest neighbor distance (in cm) between point clouds, capturing coverage and precision without requiring correspondences.
    \item \textbf{Earth Mover's Distance (EMD)} quantifies the minimum cost (in cm) of transforming one point cloud into another, providing a holistic measure of distributional similarity.
\end{itemize}

\noindent\textbf{Visual Metrics.} To evaluate action-conditioned video prediction, we render future frames. We then compare these rendered frames against the ground truth using three metrics:

\begin{itemize}[leftmargin=*,labelsep=5pt]
    \item \textbf{J-Score (IoU)} measures intersection over union of predicted and ground truth masks, quantifying spatial overlap.
    \item \textbf{F Score} evaluates contour accuracy, determining how well the boundary of the predicted object mask matches the boundary of the ground truth.
    \item \textbf{Learned Perceptual Image Patch Similarity (LPIPS)} assesses perceptual similarity using features extracted from predicted and ground truth images, capturing visual differences closer to human perception.
\end{itemize}

\subsection{Baselines}
\noindent We compare \methodshort against five baselines: two analytical physics-based simulations and two learning-based approaches.

\noindent\textbf{Spring-Mass Model~\cite{jiang2025phystwin}:} This is the same as the physics backbone we use for our model. Improvements over this baseline demonstrate that the learned residual model effectively captures the dynamics missed by the physics simulator alone.

\noindent\textbf{Material Point Method (MPM)~\cite{stomakhin2013material} :} 
MPM combines Eulerian and Lagrangian representations to handle material behaviors and topological changes. This hybrid approach discretizes materials into particles while using a background grid for computing spatial derivatives and enforcing conservation laws. MPM is particularly effective for materials exhibiting both solid and fluid-like behaviors but requires careful parameter tuning and can be computationally expensive.

\noindent\textbf{Graph Based Neural Dynamics (GBND)~\cite{zhang2024dynamic} :} 
This is a learned approach that employs Graph Neural Networks to learn object dynamics directly from data without explicit material parameters. The object is represented as a graph where message passing mechanisms propagate information between nodes to capture long range interactions.

\noindent\textbf{Particle-Grid Neural Dynamics (PGND)~\cite{zhang2025particle} :} 
This represents the state-of-the-art in learning-based methods, employing a hybrid representation of particles and spatial grids inspired by MPM to model deformable object dynamics. In this method, particles capture the object geometry while the spatial grid discretizes the 3D domain to ensure spatial continuity and improve learning efficiency.

\noindent\textbf{Differentiable (Diff.) Spring-Mass~\cite{jiang2025phystwin}:} This extends the spring-mass baseline with a two-stage optimization. It first estimates global parameters as spring-mass baseline, then refines them through gradient-based optimization using a differentiable simulator to learn spatially varying physical parameters. This baseline is motivated by PhysTwin~\cite{jiang2025phystwin}, but we implement a variant that builds one canonical spring graph and registers it to each episode's initial observations before fitting the physical parameters across episodes. While this yields a more expressive spring-mass model, the dynamics remain constrained by the underlying spring-mass formulation.

\subsection{Results}
\label{sec:results}

\noindent As explained in Section \ref{sec:tasks}, we evaluate \methodshort in two ways: 3D tracking accuracy and action-conditioned video prediction.

\begin{figure*}[t]
  \centering
  \includegraphics[width=\textwidth]{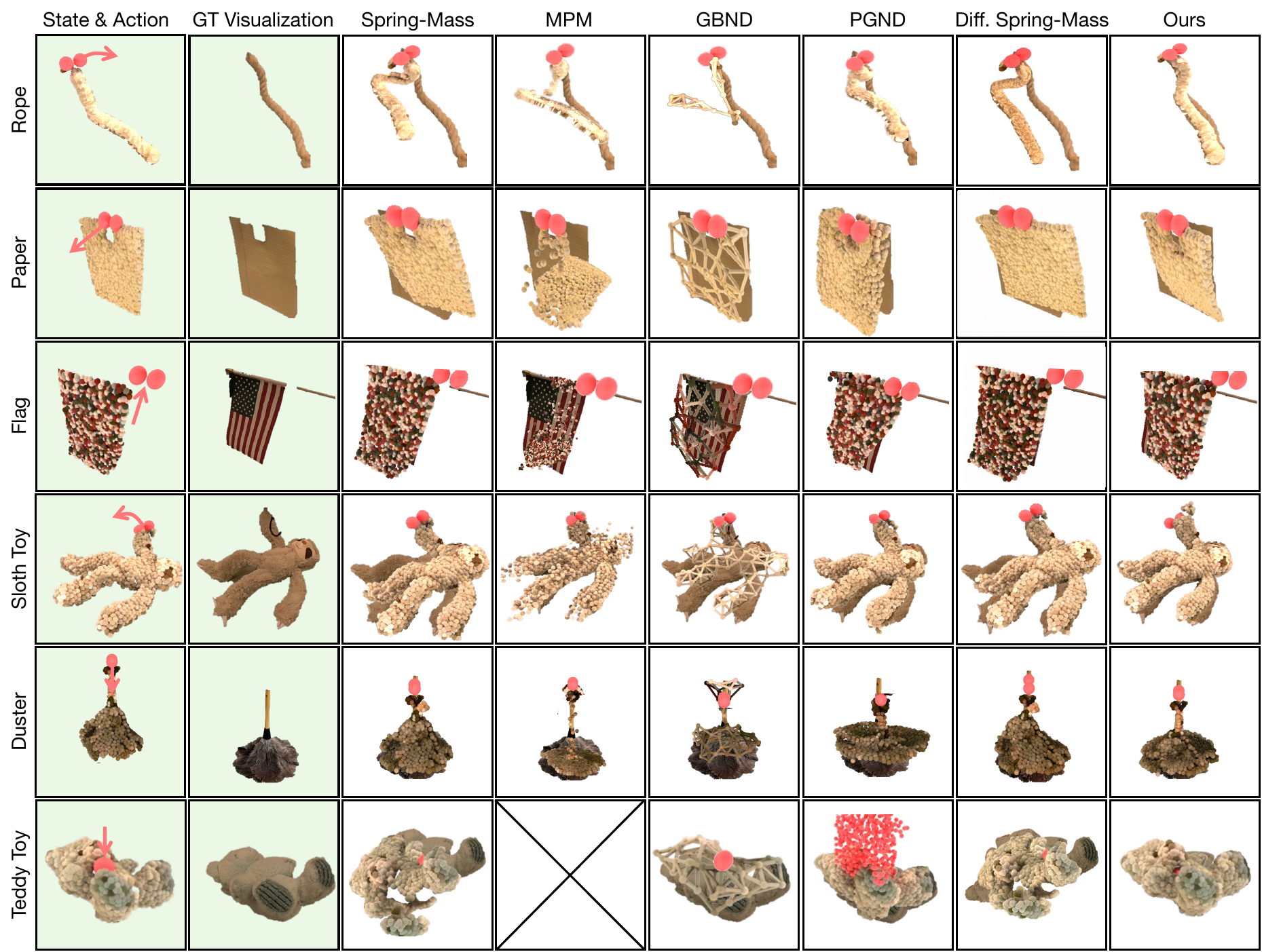}
  \vspace{-0.4cm}
  \caption{\small \textbf{Qualitative comparison of \methodshort on dynamics prediction.} Columns shaded in green show the input state with gripper positions (red spheres) and the ground truth object configuration after taking the action. The remaining columns show predictions from each method. \methodshort accurately captures deformations across all objects. For the duster, only \methodshort maintains stem rigidity while allowing deformation for the feathers. The MPM baseline is not evaluated on the Teddy Toy because it does not model the non-prehensile manipulation used in this experiment. For the Teddy Toy, PGND uses sampled gripper surface points for simulation, which are visualized as red spheres, making this example denser than the others.}
  \label{fig:qualitative}
  \vspace{-0.4cm}  %
\end{figure*}

\vspace{2mm}
\noindent \textbf{Dynamics and Tracking Accuracy.} We first analyze our model's ability to track and simulate object states over time, with results summarized in \tabref{tab:results-tracking}. 
\methodshort consistently achieves the lowest error across all objects and metrics, demonstrating that learned residual corrections on physics backbones improve accuracy beyond either pure physics or pure learning approaches.
Among physics methods, the optimized spring-mass model generally outperforms MPM, particularly for objects with coherent structure like the rope and sloth toy.
Among learning-based methods, PGND substantially outperforms GBND across all objects, indicating that particle representations with physics priors are more effective than graph neural approaches for deformable object simulation. Building on the optimized spring-mass model, Diff. Spring-Mass improves over the optimized spring-mass baseline on most objects, showing that differentiable parameter refinement provides a stronger physics-based model. However, it still falls short of \methodshort, whose learned residual corrections can capture dynamics beyond the spring-mass formulation.

Qualitatively, \figref{fig:qualitative} reveals consistent patterns in how different approaches handle deformation.
The optimized spring-mass model captures overall motion but lacks the flexibility to represent local variations in material properties, often appearing too stiff.
In our implementation, the MPM baseline often struggles to maintain structural cohesion, with object particles separating rather than behaving as connected solids.
Learning-based methods show complementary strengths: PGND generates smooth predictions but accumulates drift over time, while GBND struggles with graph topology, producing unphysical artifacts.
Diff. Spring-Mass consistently matches the observed deformation better than the spring-mass model itself, suggesting that differentiable parameter optimization improves the fitted physical response. However, because it is still constrained by the spring-mass formulation, it remains less flexible than our residual model.
\methodshort leverages its physics backbone for stability while using learned residuals to correct local dynamics, maintaining both plausibility and accuracy throughout the rollout.

The qualitative results in \figref{fig:qualitative} provide further insights into the limitations of different approaches. In our implementation, the MPM baseline performs poorly on thin objects like paper and flag, causing them to break apart and lose structure. We hypothesize that this behavior arises from the interaction between the grid discretization and the sparse particle representation used for these thin objects.
The duster, with its rigid handle and soft feathers, reveals distinct failure modes across all baselines: the optimized spring-mass model cannot assign spatially varying stiffness and causes the rigid stem to bend unnaturally; MPM fails to maintain volume with material points collapsing toward the floor; GBND predicts erroneous upward motion for all feather particles; and PGND incorrectly compresses the rigid stem. Only \methodshort simulates the duster correctly.
For the teddy, under non-prehensile manipulation, the object slips out of the optimized spring-mass model. The differentiable spring-mass model cannot accurately model the spatially varying parameters for the duster. We do not report MPM results for the Teddy Toy because the evaluated MPM baseline does not model the non-prehensile manipulation required for this task. GBND predicts minimal particle displacement, producing an overly stiff response. PGND shows the gripper's entire surface in our visualization, but the gripper fingers themselves are not visible because they have pushed into the teddy's interior. Only our \methodshort successfully captures the deformation, properly maintaining contact while allowing realistic compression.

\begin{figure*}[b]
    \centering
    \includegraphics[width=0.8\linewidth]{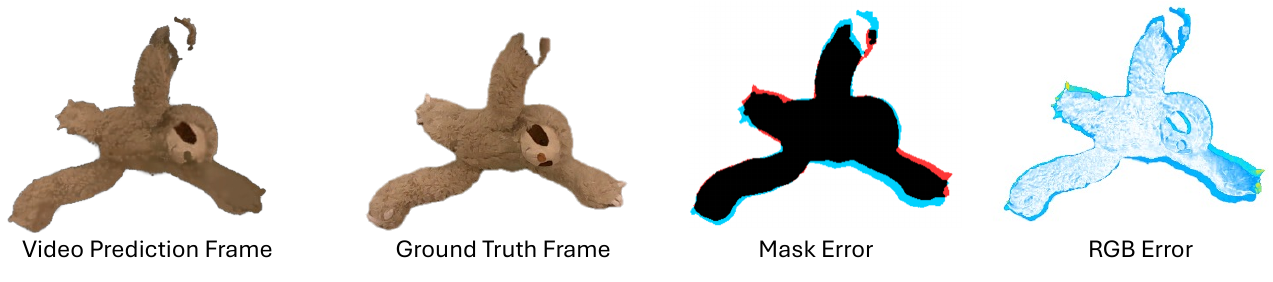}
    \vspace{-3mm}
    \caption{\small \textbf{Example pair used to compute the visual metrics.} We evaluate only the object of interest, ignoring the background. The mask error visualization shows true-positive object pixels in black, false negatives in red, and false positives in cyan. The RGB error heat map shows the per-pixel appearance difference within the evaluated object region, with stronger colors indicating larger RGB error. Part of the upper arm is missing because the sloth was grasped by the gripper, which is now removed.}
    \label{fig:video_pred}
\end{figure*}

\begin{table*}[t] 
    \centering 
    \resizebox{\textwidth}{!}{%
    \begin{tabular}{@{} l l !{\vrule width 1pt} c c c c c c @{}} 
        \thickhline 
        Method & Metric & Rope & Paper & Flag & Sloth Toy & Duster & Teddy Toy\\ 
        \thickhline 
        
        Spring-Mass~\cite{jiang2025phystwin} & \multirow{6}{*}{$\mathcal{J}$\text{-}Score / IoU ($\times 10$) $\uparrow$}
        & $3.17_{\pm 1.51}$
        & $6.82_{\pm 1.47}$ 
        & $4.60_{\pm 1.95}$ 
        & $5.30_{\pm 0.87}$ 
        & $6.66_{\pm 0.82}$ 
        & $5.43_{\pm 1.99}$ \\ 

        MPM~\cite{stomakhin2013material} & 
        & $2.44_{\pm 1.45}$ 
        & $4.51_{\pm 0.94}$ 
        & $3.78_{\pm 1.12}$
        & $5.22_{\pm 0.88}$ 
        & $5.92_{\pm 0.87}$ 
        & -- \\ 
        
        GBND~\cite{zhang2024dynamic} & 
        & $0.53_{\pm 0.55}$ 
        & $5.57_{\pm 1.40}$ 
        & $5.08_{\pm 1.42}$ 
        & $3.70_{\pm 1.49}$ 
        & $5.13_{\pm 1.04}$ 
        & $7.52_{\pm 0.66}$ \\ 
        
        PGND~\cite{zhang2025particle} & 
        & $3.18_{\pm 2.13}$  
        & $7.08_{\pm 0.90}$ 
        & $5.47_{\pm 1.66}$ 
        & $5.86_{\pm 0.93}$ 
        & $6.69_{\pm 0.46}$ 
        & $7.13_{\pm 1.02}$ \\ 
        
        Diff. Spring-Mass~\cite{jiang2025phystwin} &
        & $3.45_{\pm 1.43}$
        & $7.10_{\pm 1.37}$
        & $4.98_{\pm 1.82}$
        & $5.58_{\pm 0.82}$
        & $6.70_{\pm 0.78}$
        & $5.82_{\pm 1.85}$ \\

        Ours & 
        & $\mathbf{3.95_{\pm 1.69}}$
        & $\mathbf{7.53_{\pm 0.82}}$ 
        & $\mathbf{5.54_{\pm 1.89}}$ 
        & $\mathbf{6.21_{\pm 0.78}}$ 
        & $\mathbf{6.77_{\pm 0.83}}$ 
        & $\mathbf{7.82_{\pm 1.14}}$ \\ 
        \thickhline 
        
        Spring-Mass~\cite{jiang2025phystwin} & \multirow{6}{*}{$\mathcal{F}$\text{-}Score ($\times 10$) $\uparrow$}
        & $6.59_{\pm 1.60}$
        & $4.71_{\pm 2.61}$ 
        & $2.08_{\pm 1.81}$ 
        & $5.76_{\pm 1.12}$ 
        & $5.32_{\pm 0.97}$ 
        & $4.73_{\pm 2.72}$ \\ 

        MPM~\cite{stomakhin2013material} & 
        & $4.92_{\pm 2.45}$ 
        & $4.73_{\pm 0.82}$ 
        & $2.49_{\pm 0.97}$
        & $5.13_{\pm 0.86}$ 
        & $5.00_{\pm 1.08}$ 
        & -- \\ 
        
        GBND~\cite{zhang2024dynamic} & 
        & $1.95_{\pm 1.91}$ 
        & $4.95_{\pm 1.35}$ 
        & $2.54_{\pm 2.04}$ 
        & $3.59_{\pm 1.15}$ 
        & $4.11_{\pm 1.09}$ 
        & $7.71_{\pm 0.96}$ \\ 
        
        PGND~\cite{zhang2025particle} & 
        & $5.87_{\pm 2.96}$  
        & $5.39_{\pm 1.65}$ 
        & $2.65_{\pm 1.98}$ 
        & $5.53_{\pm 1.11}$ 
        & $5.29_{\pm 0.97}$ 
        & $7.27_{\pm 1.54}$ \\ 
        
        Diff. Spring-Mass~\cite{jiang2025phystwin} &
        & $6.95_{\pm 1.50}$
        & $4.98_{\pm 2.45}$
        & $2.22_{\pm 1.70}$
        & $6.03_{\pm 1.05}$
        & $5.45_{\pm 0.91}$
        & $5.05_{\pm 2.50}$ \\

        Ours & 
        & $\mathbf{7.38_{\pm 1.85}}$ 
        & $\mathbf{5.91_{\pm 1.76}}$ 
        & $\mathbf{2.68_{\pm 2.14}}$ 
        & $\mathbf{6.30_{\pm 1.07}}$ 
        & $\mathbf{5.51_{\pm 1.02}}$ 
        & $\mathbf{8.03_{\pm 1.75}}$ \\ 
        \thickhline 
        
        Spring-Mass~\cite{jiang2025phystwin} & \multirow{6}{*}{LPIPS ($\times 100$) $\downarrow$}
        & $2.6_{\pm 0.9}$ 
        & $5.6_{\pm 2.2}$ 
        & $9.9_{\pm 3.0}$ 
        & $5.9_{\pm 2.0}$ 
        & $7.0_{\pm 1.9}$ 
        & $3.6_{\pm 1.7}$ \\ 

        MPM~\cite{stomakhin2013material} & 
        & $3.8_{\pm 1.4}$ 
        & $6.5_{\pm 1.4}$ 
        & $9.1_{\pm 2.5}$
        & $7.7_{\pm 1.7}$ 
        & $6.8_{\pm 1.7}$ 
        & -- \\ 
        
        GBND~\cite{zhang2024dynamic} & 
        & $5.2_{\pm 1.7}$ 
        & $5.7_{\pm 1.3}$ 
        & $8.8_{\pm 1.3}$ 
        & $10.5_{\pm 2.0}$ 
        & $6.4_{\pm 1.3}$ 
        & $2.2_{\pm 0.5}$ \\ 
        
        PGND~\cite{zhang2025particle} & 
        & $2.6_{\pm 1.3}$     
        & $5.2_{\pm 1.1}$ 
        & $8.9_{\pm 1.9}$ 
        & $6.5_{\pm 1.8}$ 
        & $6.6_{\pm 0.9}$ 
        & $2.0_{\pm 0.8}$ \\ 
        
        Diff. Spring-Mass~\cite{jiang2025phystwin} &
        & $2.35_{\pm 0.85}$
        & $5.05_{\pm 2.00}$
        & $8.95_{\pm 2.70}$
        & $5.45_{\pm 1.85}$
        & $6.35_{\pm 1.75}$
        & $3.25_{\pm 1.55}$ \\

        Ours & 
        & $\mathbf{2.3_{\pm 1.1}}$ 
        & $\mathbf{4.9_{\pm 1.4}}$ 
        & $\mathbf{8.5_{\pm 2.9}}$ 
        & $\mathbf{5.4_{\pm 2.1}}$ 
        & $\mathbf{5.9_{\pm 1.9}}$ 
        & $\mathbf{1.7_{\pm 0.9}}$ \\ 
        \thickhline 
        
    \end{tabular} } 
    \caption{\small \textbf{Action-conditioned video prediction with dynamics prediction and 3DGS.} \methodshort produces the most accurate visual predictions, demonstrating superior rendering quality.} 
    \label{tab:results-video-pred} 
    \vspace{-1em}
\end{table*}

\begin{figure*}[t]
  \centering
  \includegraphics[width=\linewidth]{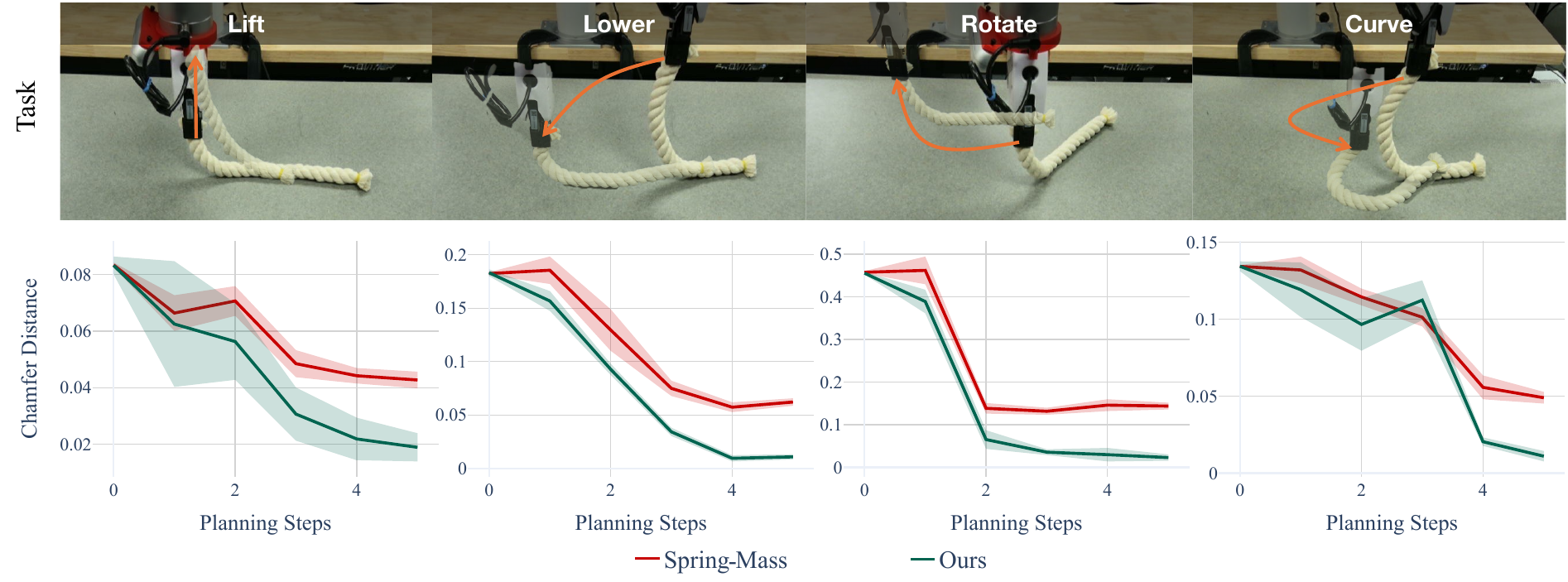}
  \vspace{-0.55cm}
  \caption{\small \textbf{Planning results on rope manipulation tasks.} Top row shows start and end configurations for four tasks. The bottom row shows the Chamfer Distance to the target during MPPI planning. \methodshort (green) converges faster and achieves lower final error than the optimized spring-mass baseline (red) across all tasks.}
  \label{fig:rope_results}
  \vspace{-0.2cm}  %
\end{figure*}

\vspace{2mm}
\noindent \textbf{Action Conditioned Video Prediction.}
Given an initial observation and a sequence of actions, we evaluate whether accurate dynamics predictions translate to realistic future frame generation.
We use 3D Gaussian Splatting as the rendering module, where Gaussians are fit to the observed point cloud at the initial timestep and parameterized by position, covariance, opacity, and color. As \methodshort predicts future point cloud configurations, we update the Gaussian positions and covariances while keeping opacity and color fixed, allowing the rendered frames to reflect the predicted deformations while preserving appearance consistency with the initial observation. At each timestep, we project the updated Gaussians into the camera view and alpha-blend them in depth order to produce the output image.
\figref{fig:video_pred} shows an example pair consisting of a frame rendered from the dynamics predicted by \methodshort and the corresponding ground truth frame, which are used to compute the visual metrics. Among the physics-based methods, Diff. Spring-Mass generally improves over the optimized spring-mass baseline, indicating that differentiable parameter refinement translates to more accurate rendered masks and appearances. Compared with the learning-based method PGND, its performance is mixed: it is better on some objects and metrics, but worse on others. Overall, \methodshort consistently achieves the best $\mathcal{J}$-Score and $\mathcal{F}$-Score across all objects, demonstrating better spatial overlap and foreground detection compared to both physics-based and learning-based baselines. \methodshort also maintains the lowest LPIPS scores, indicating that the physics backbone enables the residual model to preserve object appearance during rendering.

\section{Applications} \label{sec:applications}

\begin{figure}[t]
  \centering
  \includegraphics[width=\linewidth]{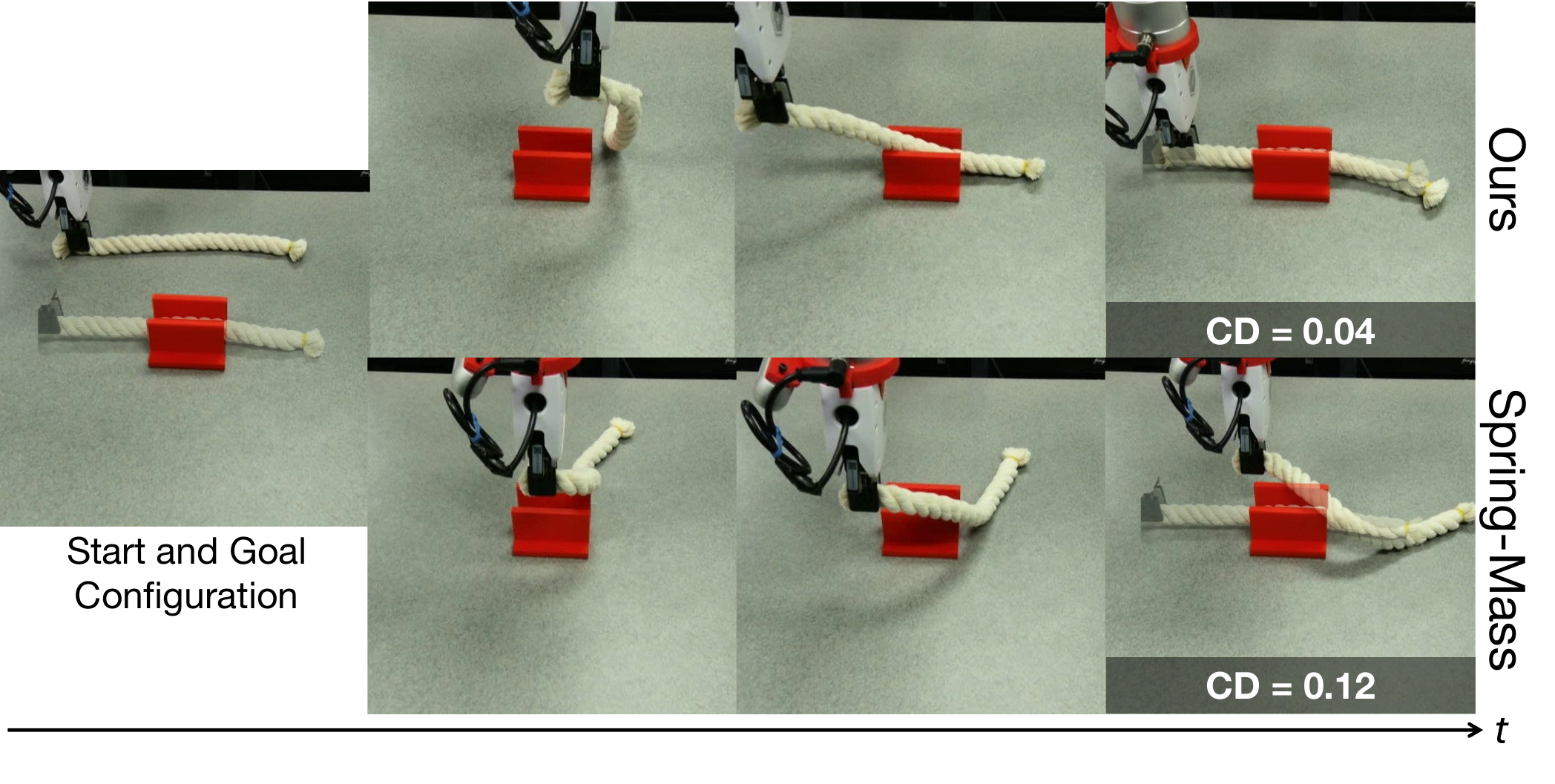}
  \vspace{-0.8cm}  %
  \caption{\small \textbf{Cable Rerouting Execution.} Comparison of \methodshort (top) and the Spring-Mass backbone (bottom) for rope rerouting through a slot. \methodshort navigates the rope through the opening while the Spring-Mass backbone exhibits repeated collisions with the slot edges.}

  \label{fig:rerouting_execution}
  \vspace{-0.4cm}  %
\end{figure}

\subsection{Planning using \methodshort}
\label{sec:planning}
In our planning experiments, we use \methodshort as a forward model for model-based control: given a state $s_t$ and an action $a_t$, it predicts the next state $s_{t+1}$, where actions specify gripper positions that interact with the object. To plan over this forward model, we integrate \methodshort with Model Predictive Path Integral (MPPI) control~\cite{garcia1989model}. At each timestep, MPPI samples a batch of candidate action sequences $\{\tau_i\}$ spanning a planning horizon $H$, rolls out \methodshort to predict future states, and updates the action distribution toward lower-cost regions. Following standard MPC practice, only the first action is executed before replanning with the newly observed state. We use the Chamfer Distance (CD) between the predicted and a pre-collected target point cloud as the planning objective. The parameters for planning experiments is provided in \appref{app:mppi_parameters}.

\vspace{2mm}
\noindent \textbf{Standard Manipulation Tasks.}
We evaluate planning on the rope with four start and goal configurations, as shown in \figref{fig:rope_results} (top): (1) lifting the rope up, (2) lowering it, (3) rotating it such that one end remains roughly stationary while the other moves, and (4) deforming it into a curved shape. We compare \methodshort against the spring-mass backbone, providing both methods with the same MPPI planning budget. We run 10 trials for each task using the same start and end configurations. As shown in \figref{fig:rope_results}, \methodshort's CD decreases smoothly and saturates at a value close to the target, whereas the spring-mass model converges more slowly and sometimes fails to reach a comparable final error.  Additional results on the sloth toy are provided in \appref{app:plush_planning}.

\begin{figure}[t]
  \centering
  \includegraphics[width=\linewidth]{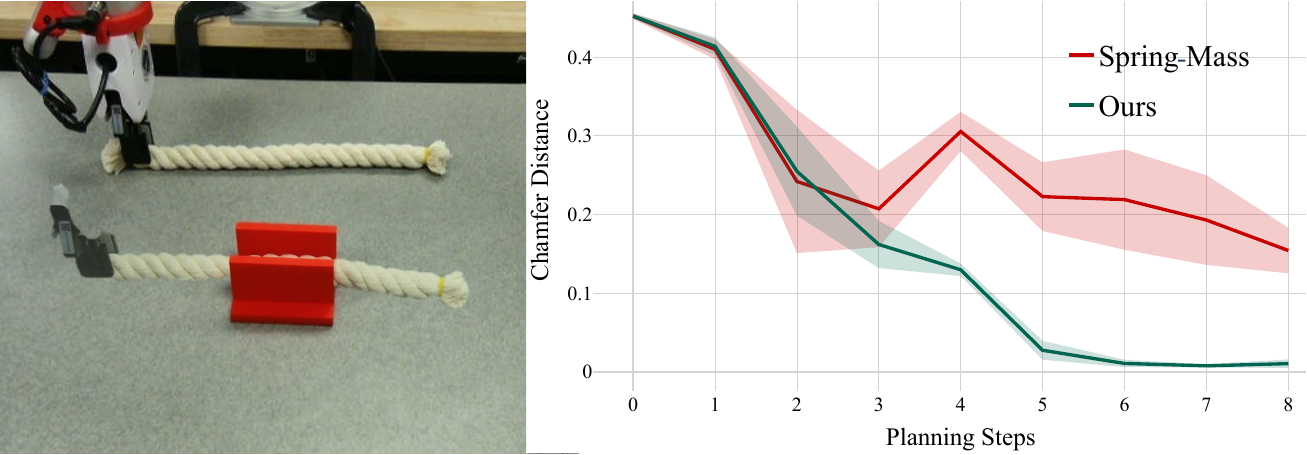}
  \vspace{-0.6cm}  %
  \caption{\small \textbf{Cable rerouting through narrow slot.} Left: distribution of start and end configurations. Right: Chamfer Distance during MPPI planning steps. \methodshort achieves lower distance than the spring-mass baseline, successfully threading the rope through the slot in 8 out of 10 trials compared to 2 out of 10 for the baseline.}
  \label{fig:rerouting}
  \vspace{-0.4cm}  %
\end{figure}

\begin{figure*}[t]
  \centering
  \includegraphics[width=\linewidth]{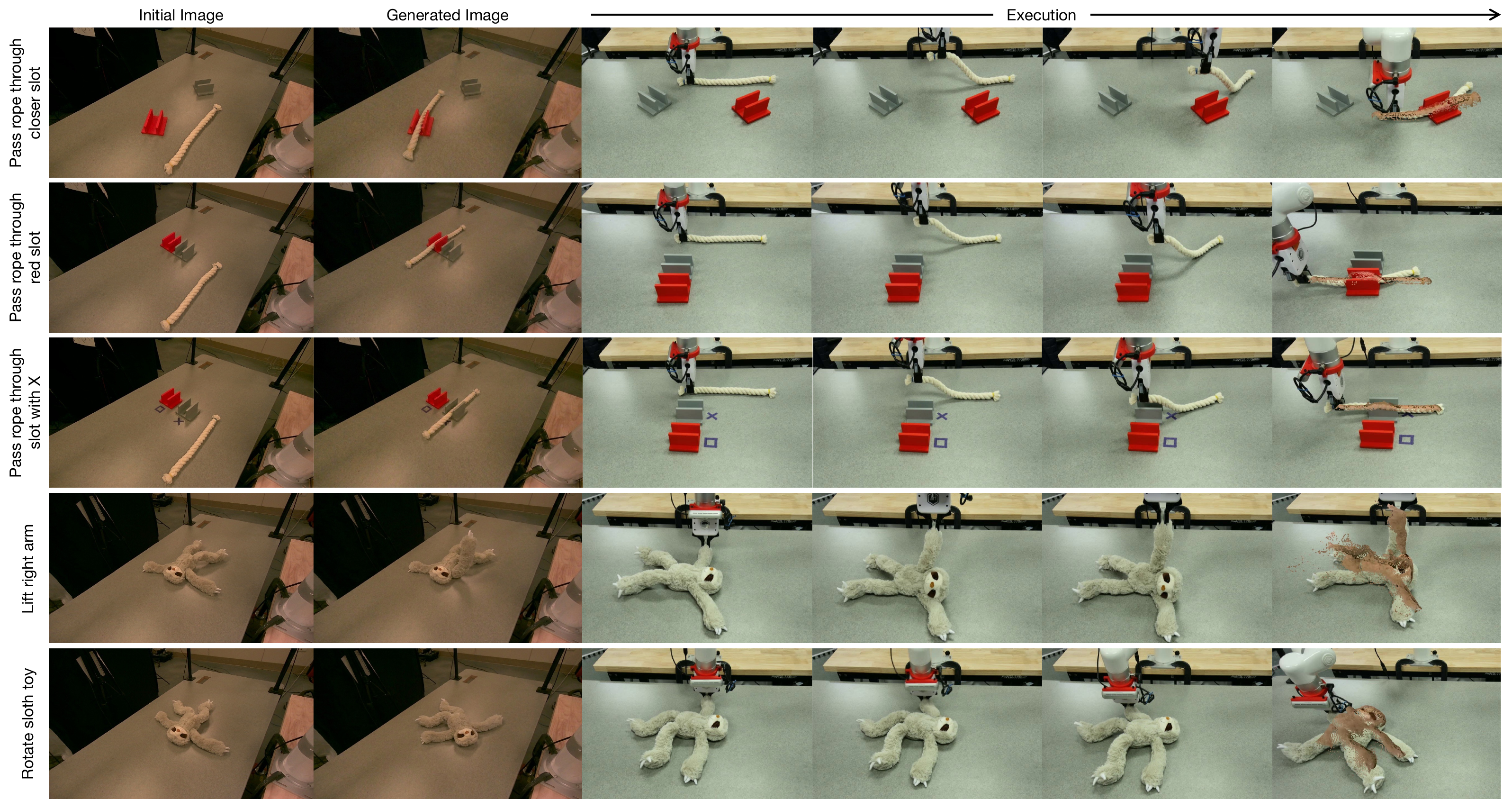}
  \vspace{-0.4cm}
  \caption{\small \textbf{Execution of \methodshort with generated goals.} The first
  column shows the initial image, and the second column shows the language-conditioned generated goal image. Subsequent columns show robot execution from a higher-quality demo camera. The target point cloud from the goal image is overlaid on the last image. \methodshort successfully executes a range of tasks, including passing a rope through a specified slot and moving a sloth toy to target configurations.}
  \label{fig:demos}
  \vspace{-0.2cm}
\end{figure*}

\vspace{2mm}
\noindent \textbf{Cable Rerouting.}
We further test \methodshort on a more challenging task: cable rerouting through a narrow slot. This requires the robot to move the rope and thread it through the slot. For these experiments, the residual dynamics model is not trained on examples containing rope-slot interaction data. Hence, this also shows the generalization capability of \methodshort. We run 10 trials with different start and end configurations. The target configuration is provided to the planner.

As demonstrated in \figref{fig:rerouting_execution}, \methodshort successfully navigates the rope through the slot. The learned residuals compensate for inaccuracies in the physics backbone, enabling precise trajectory execution that avoids collisions with the edges. In contrast, the spring-mass backbone struggles with these fine dynamics. Small errors cause the rope to collide with the edges, forcing the planner to retry the motion multiple times before completing the task.

Quantitatively, \figref{fig:rerouting} shows the distributions of start and end configurations overlaid (left) and comparisons of chamfer distances (right). \methodshort achieves significantly lower CD than the optimized spring-mass. We also evaluate success rates for this task. A trial is considered successful if the rope passes entirely through the slot and the gripper remains below the slot height in the final configuration. Under this criterion, \methodshort succeeds in 8/10 trials, whereas the spring-mass model succeeds in only 2/10 trials. These results demonstrate that the improved model enables more reliable execution of challenging manipulation tasks.

\vspace{2mm}
\noindent \textbf{Language-Conditioned Planning.}
The planning experiments above require a target point cloud, so it must be collected beforehand. To relax this requirement, we integrate \methodshort with a language-conditioned goal-generation pipeline, enabling the planner to operate solely on language commands.

Given an initial RGB image and a language command (e.g., ``pass rope through red slot''), we use Nano Banana Pro~\cite{google_nanobanana_pro_2025} to generate a goal image. Note that here we use a single camera rather than the four in the standard planning experiments, since the goal image can be generated more reliably for a single viewpoint. To estimate depth from the generated image, we use Depth Anything V2~\cite{yang2024depth}. Following Patel \textit{et al.}~\cite{patel2025robotic}, we resolve the scale and shift ambiguity inherent to monocular depth estimation by fitting an affine transform to align the predicted depth to the initial depth map from the depth camera. We restrict the alignment to background table points, which remain consistent in both images. Since the object appears in a different configuration in the goal image than in the initial image, including object points for this alignment would degrade the quality. The resulting depth map is unprojected into 3D with known camera parameters to obtain the target point cloud, which is passed directly to the MPPI planner. 

To determine where to grasp the object, we use AnyGrasp~\cite{fang2023anygrasp} to generate grasp candidates on the initial point cloud. For the rope, we select the candidate closest to the right end. For the sloth toy, the language command implies a specific region to manipulate, which we identify by extracting DINOv2~\cite{oquab2023dinov2} features from both the initial and goal images to establish dense correspondences. We locate the object point with the largest displacement between the two images and select the AnyGrasp candidate closest to that point.

\figref{fig:demos} shows real-world execution for multiple tasks. Beyond removing the need to physically collect the goal configuration, this pipeline opens up a richer space of goals, including spatial references such as ``pass rope through closer slot'' and ``lift right arm'' of the sloth toy.

\begin{figure}[htbp]
    \centering
    \vspace{-0.2cm}
    \includegraphics[width=\columnwidth]{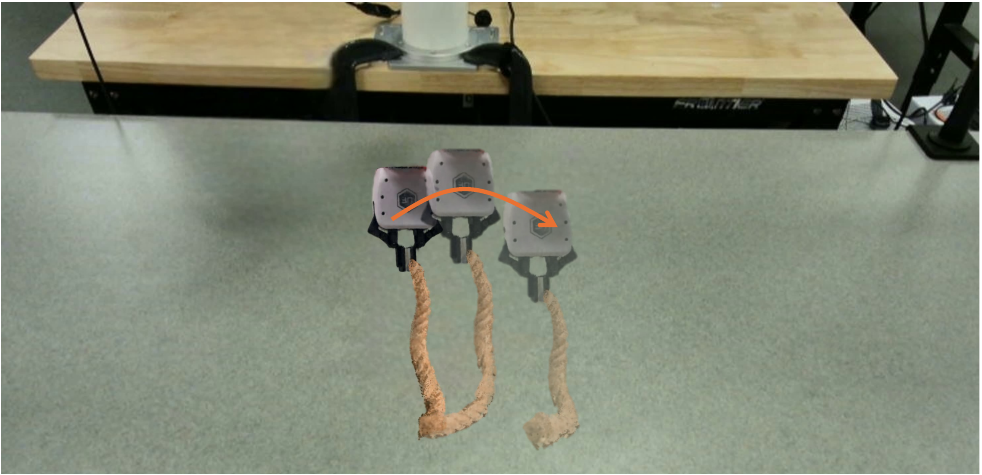}
    \caption{\small \textbf{Interactive rope manipulation with \methodshort rendered using 3DGS.} The visualization overlays three timesteps, with later configurations in lighter opacity. As \methodshort predicts the dynamics, the 3D Gaussians attached to the particles are updated to render images that maintain visual consistency.}
    \label{fig:interactive_rope}
    \vspace{-0.3cm}
\end{figure}

\subsection{Interactive Photo-Realistic Simulation.}

As shown in \figref{fig:interactive_rope}, \methodshort supports interactive, photo-realistic simulation by combining its predicted dynamics with 3D Gaussian Splatting.
Users issue manipulation commands via keyboard input; at each step, the physics backbone propagates the object state, the learned residual network applies per-particle corrections, and the updated Gaussians are rendered to produce a photo-realistic view of the resulting deformation.
Because only Gaussian positions are updated while appearance attributes remain fixed, the rendering maintains visual consistency with the initial observation throughout the interaction.

\section{Conclusions}
This paper presented  Physics-Guided Residual Dynamics (\methodshort), a hybrid simulation framework that bridges physics-based and learning-based approaches for deformable object simulation. By combining an optimized spring-mass backbone with learned velocity residuals, \methodshort achieves superior accuracy while maintaining physical plausibility. Our velocity-based residuals, combined with a sliding-window transformer for temporal aggregation, yield stable predictions, avoiding the instability issues plaguing naive residual learning. Extensive experiments across real-world objects demonstrate that \methodshort consistently outperforms a variety of SOTA methods. \methodshort handles challenging scenarios involving heterogeneous materials, volumetric deformations, and contact-rich interactions where existing methods struggle. Beyond tracking, \methodshort also shows promise for applications like action-conditioned video prediction, planning, and interactive simulation. %

\bibliographystyle{plainnat}
\bibliography{references}

\clearpage
\renewcommand{\thesection}{\Alph{section}}
\renewcommand{\theHsection}{appendix.\Alph{section}}  %
\setcounter{section}{0}
\setcounter{page}{1}
\maketitlesupplementary

\noindent We structure the supplement into the following sections:
\begin{itemize}[leftmargin=1.5em, label={}]
    \item[\textcolor{blue}{\hyperref[sec:ablation_appendix]{\ref*{sec:ablation_appendix}}}] Ablation study of the main components of \methodshort, including the encoder, decoder, temporal aggregator, gating, and internal particles for volumetric objects.
    \item[\textcolor{blue}{\hyperref[app:network_details]{\ref*{app:network_details}}}] Network architecture details and hyperparameter settings for the encoder, decoder, and temporal aggregator.
    \item[\textcolor{blue}{\hyperref[app:plush_planning]{\ref*{app:plush_planning}}}] Planning experiments on the sloth toy, showing \methodshort's behavior on a volumetric object with long flexible limbs.
    \item[\textcolor{blue}{\hyperref[app:data_coll]{\ref*{app:data_coll}}}] Data collection and processing pipeline for extracting temporally consistent particle trajectories from multi-view RGBD observations.
    \item[\textcolor{blue}{\hyperref[app:mppi_parameters]{\ref*{app:mppi_parameters}}}] MPPI planning hyperparameters used for the simple tasks and the cable rerouting experiments.
\end{itemize}

\section{Ablation Study}
\label{sec:ablation_appendix}

\begin{table}[h]
    \centering
    \small
    \setlength{\tabcolsep}{5pt}
    \resizebox{0.8\columnwidth}{!}{%
        \begin{tabular}{lcc}
            \toprule
            Ablation variant & Rope & Sloth Toy \\
            \midrule
            \textbf{\methodshort} & \textbf{2.6} & \textbf{2.7} \\
            Replace PTv3 with PointNet++ & 3.0 & 3.2 \\
            Remove decoder & 2.8 & 3.0 \\
            Remove temporal aggregator & 2.8 & 3.1 \\
            Replace gating with addition & 2.7 & 2.9 \\
            Remove internal particles & N/A & 3.8 \\
            \bottomrule
        \end{tabular}%
    }
    \caption{\small \textbf{Ablations of the main components of our model.} We report MDE. Replacing velocity with position residuals leads to divergence. Rope has no internal particles, so it is N/A.}
    \label{tab:abl_rebut}
    \vspace{-3mm}
\end{table}

We provide a detailed ablation study of the main components of \methodshort. We consider both replacement ablations, which test natural alternatives to our design, and removal ablations, which isolate the role of individual components. All quantitative results are summarized in \tabref{tab:abl_rebut}. We report only MDE, and the other metrics follow the same trends.

\subsection{Encoder choice}
\label{sec:ablation_encoder}
\noindent  We use PTv3 as our encoder. We compare this design to a PointNet++~\cite{qi2017pointnet++}, which is a natural baseline because it is a standard point cloud architecture and also used in PGND. This comparison tests whether the stronger local feature extraction of PTv3 is important for residual prediction. The trend in \tabref{tab:abl_rebut} indicates that it is.

\subsection{Decoder}
\label{sec:ablation_decoder}
\noindent In \secref{sec:network_arch}, we describe a NeRF-style decoder that combines encoder features with positional encoding to predict per-particle residual velocities. A natural simplification is to remove this decoder and directly project the encoder features to residual velocities. This tests whether explicit position-conditioned decoding remains useful once the encoder features are already available. The results in \tabref{tab:abl_rebut} suggest that it does.

\subsection{Temporal aggregator}
\label{sec:ablation_temporal}
\noindent The temporal aggregator refines the current residual estimate using a sliding window of previous predictions. This design is motivated by the fact that many deformation effects depend on recent motion history rather than on the current configuration alone. The corresponding ablation removes temporal refinement while keeping the encoder and decoder unchanged. The degradation in \tabref{tab:abl_rebut} supports the need for temporal context in predicting residual corrections.

\subsection{Gated temporal refinement}
\label{sec:ablation_gating}
\noindent Our temporal module uses gating to blend the decoder prediction with the temporally refined correction. A natural alternative is direct addition. This ablation isolates whether the gate is important beyond the temporal model itself. As shown in \tabref{tab:abl_rebut}, the gate is useful for turning temporal information into a stable refinement rather than an uncontrolled update.

\subsection{Internal particles for volumetric objects}
\label{sec:ablation_internal_particles}
\noindent For volumetric objects, the spring mass backbone includes internal particles so the particles don't collapse. This component is relevant only to the sloth and the teddy toy. The corresponding ablation removes the internal particles while keeping the rest of the system unchanged. The resulting drop in \tabref{tab:abl_rebut} is consistent with the role of internal particles in preventing unrealistic collapse for volumetric objects.

\section{Network Architecture Details}
\label{app:network_details}

Our network network consists of three primary components: an encoder, a decoder, and a temporal aggregator. The hyperparameter values for each component are provided in Table~\ref{tab:network_hyperparams}. The encoder uses a Point Transformer V3 (PTv3)~\cite{wu2024point} architecture to extract per-particle features through patch-based self-attention, where the feature dimension determines the size of the latent representation. The input channels correspond to the concatenated position and velocity information from the current state, history timesteps, and the physics simulator predictions. The decoder employs a conditional NeRF-style MLP to map spatial locations to residual velocities, where hidden layers refers to the number of fully connected layers, hidden dimension specifies the width of each layer, and output channels indicates the dimensionality of the predicted residual velocity (3 for xyz components). The decoder takes as input the per-particle features from the encoder, concatenated with Fourier positional encodings of particle positions. The temporal aggregator employs a transformer encoder to refine predictions using temporal context. Here, the embedding dimension defines the size of the latent space for temporal features, attention heads specifies the number of parallel attention mechanisms, transformer layers indicates the depth of the transformer stack, and feedforward dimension determines the width of the intermediate feedforward network within each transformer layer. Finally, the gating scale controls the magnitude of temporal corrections applied to the base velocity predictions, preventing large adjustments that could destabilize the simulation.

\begin{table}[h]
\centering
\begin{tabular}{lc}
\toprule
\textbf{Parameter} & \textbf{Value} \\
\midrule
\multicolumn{2}{l}{\textit{Encoder (PTv3)}} \\
Feature dimension & 64 \\
Input channels & 18 \\
\midrule
\multicolumn{2}{l}{\textit{Decoder}} \\
Hidden layers & 2 \\
Hidden dimension & 64 \\
Output channels & 3 \\
\midrule
\multicolumn{2}{l}{\textit{Temporal Aggregator}} \\
Embedding dimension (\(d\)) & 64 \\
Attention heads (\(H\)) & 4 \\
Transformer layers (\(L\)) & 2 \\
Feedforward dimension (\(d_{\text{ff}}\)) & 128 \\
Gating scale & 0.1 \\
\bottomrule
\end{tabular}
\caption{Network architecture hyperparameters.}
\vspace{-5mm}
\label{tab:network_hyperparams}
\end{table}

\begin{figure*}[t]
  \centering
  \includegraphics[width=\linewidth]{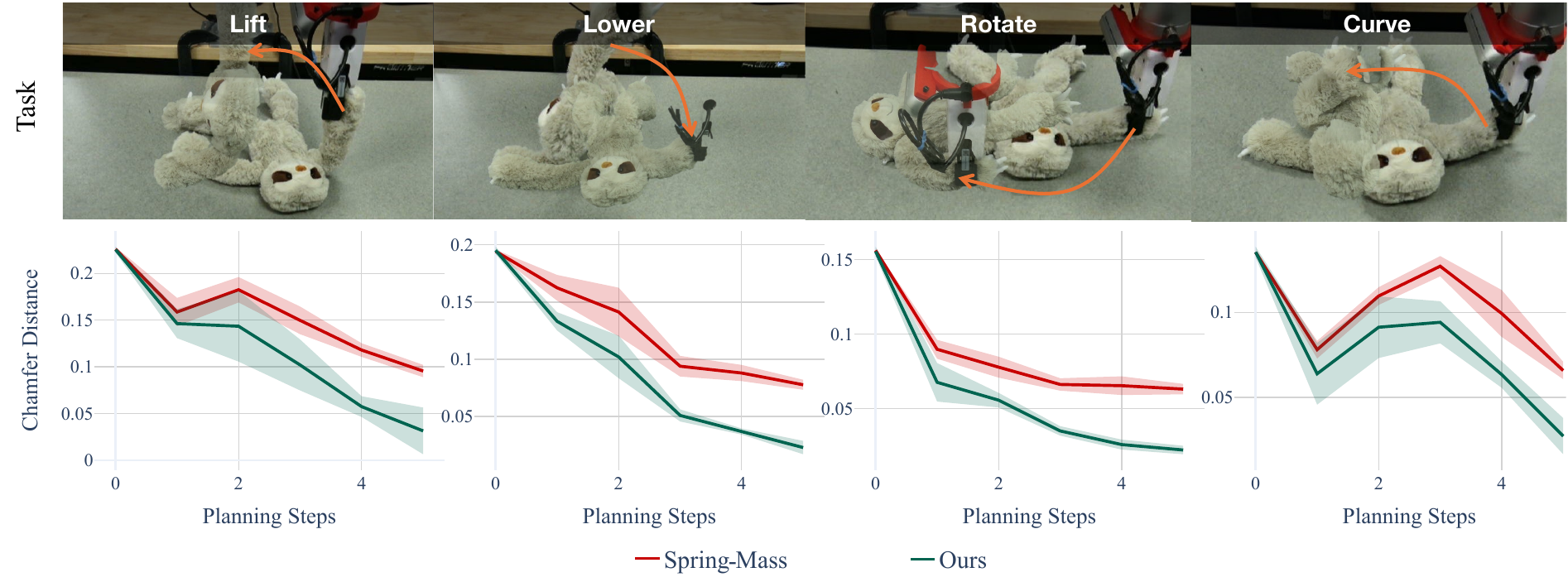}
  \vspace{-0.55cm}
  \caption{\small \textbf{Planning results on sloth toy manipulation tasks.} Top row displays the initial and target configurations for four manipulation objectives. Bottom row plots the Chamfer Distance evolution throughout MPPI planning. \methodshort (green) consistently achieves superior convergence compared to the tuned spring-mass baseline (red).}
  \label{fig:sloth_results}
  \vspace{-0.6cm}
\end{figure*}

\section{Planning on Sloth Toy}
\label{app:plush_planning}
We extend our planning experiments to the sloth toy, a volumetric object with long, flexible limbs. It requires accurate prediction of both volumetric deformation and limb dynamics, testing the framework's ability to handle three-dimensional objects. Similar to rope, we design four manipulation objectives with varying complexity, as visualized in \figref{fig:sloth_results} (top): (1) lifting the object, (2) lowering it, (3) rotating it, and (4) bending it into a curved configuration. Following the same experimental setup, we conduct 10 trials per objective with identical initial and target states, allocating equal computational budget to both \methodshort and the spring-mass baseline. The results in \figref{fig:sloth_results} demonstrate that \methodshort exhibits consistent convergence behavior, reaching target configurations with reasonable residual error. The spring-mass baseline shows slower error reduction and occasionally plateaus at suboptimal solutions, particularly for tasks involving complex limb deformation, where the learned residuals provide crucial corrections to the physics backbone predictions.

\section{Data Collection and Processing}
\label{app:data_coll}
We mount four Intel RealSense D455 cameras around the workspace to capture synchronized RGBD observations at 30 Hz. The cameras are calibrated to a shared world coordinate frame using a checkerboard calibration procedure, enabling the fusion of observations across views. For each camera view, we apply Grounded SAM 2 to extract object masks. Using text prompts containing object descriptions, it detects and segments the object in the first frame of a sequence and propagates the mask across subsequent frames. We employ CoTracker~\cite{karaev2025cotracker3} as the tracking model due to its stable tracking performance. It predicts a set of 2D trajectories, initialized from uniformly sampled grid locations within the first-frame object mask. For each pixel in the segmented region at time $t$, we compute its 2D displacement to time $t+1$ from the predicted trajectories, and convert it to a 2D velocity by dividing by the time step.

For each camera view, we lift the 2D pixel velocities to 3D using the corresponding depth measurements. Specifically, for a pixel at position $(u, v)$ with depth $d$ and 2D velocity $(v_u, v_v)$, we compute the 3D velocity by back-projecting both the current and next pixel positions to 3D coordinates using camera intrinsics, then computing the difference. This yields per-pixel 3D velocities in the camera frame, which we transform to the world frame using the calibration parameters. We aggregate velocities across all camera views by averaging the 3D velocity estimates for overlapping spatial regions, thereby improving robustness to noise and partial observations from individual views.

Given the 3D point cloud with per-point velocities at each timestep, we extract temporally consistent particle trajectories using an iterative rollout procedure. Starting from frame 0, we initialize particles at the point cloud positions $\{x_i^0\}$. For each subsequent timestep $t$, we propagate particles forward using their current velocities: $x_i^{t+1} = x_i^t + v_i^t \cdot \Delta t$. To update velocities at the new positions, we perform k-nearest neighbor search (with $k=5$) between the propagated particle positions and the observed point cloud at time $t+1$. Each particle's velocity is updated to the mean velocity of its k nearest neighbors in the observed cloud. This iterative process maintains correspondence across frames while being robust to tracking noise, yielding persistent point tracks suitable for training our dynamics model. The entire processing pipeline runs at approximately 2 Hz for sequences with 1000 particles.

\section{MPPI Planning Parameters}
\label{app:mppi_parameters}
We use different set of parameters for simple and cable rerouting experiments. They are provided in \tabref{tab:mppi_params}.

\begin{table}[h]
    \centering
    \small
    \resizebox{\columnwidth}{!}{
    \begin{tabular}{lcc}
        \toprule
        \textbf{MPPI Hyperparameter} & \textbf{Simple tasks} & \textbf{Cable rerouting} \\
        \midrule
        Number of sampled trajectories & 8 & 15 \\
        Planning horizon length & 10 & 15 \\
        Optimization iterations per step & 3 & 6 \\
        Action noise covariance & 0.1 & 0.1 \\
        \bottomrule
    \end{tabular}
    }
    \caption{\small \textbf{MPPI hyperparameters used for planning.} Separate settings are used for the simpler tasks and for cable rerouting.}
    \label{tab:mppi_params}
\end{table}

\end{document}